\documentclass[journal]{IEEEtran}
\usepackage{amsmath,graphicx}
\usepackage{color,hyperref}
\usepackage[table,xcdraw]{xcolor}

\usepackage{times}
\usepackage{epsfig}
\usepackage{graphicx}
\usepackage{amsmath}
\usepackage{amssymb}

\usepackage{algorithm}
\usepackage{algorithmicx}
\usepackage{algpseudocode}
\usepackage{multirow}
\usepackage{booktabs} 
\usepackage{url}
\usepackage{cite}
\usepackage{bm}
\usepackage{booktabs}
\usepackage{makecell}
\usepackage{comment}
\usepackage{xcolor}

\definecolor{red}{RGB}{255,0,0}
\definecolor{blue}{RGB}{68,114,196}
\definecolor{green}{RGB}{112,173,71}

\newcommand{\MSA}{\text{MSA}}
\newcommand{\MLP}{\text{MLP}}
\newcommand{\LN}{\text{LN}}

\newlength\savedwidth

\usepackage{hyperref}
\hypersetup{
    colorlinks=true,
    linkcolor=blue,
    filecolor=blue,
    urlcolor=blue
    }

\begin{document}
%
\title{RRSIS: Referring Remote Sensing Image Segmentation}
\author{Zhenghang Yuan,~\IEEEmembership{Student Member,~IEEE,} Lichao Mou, Yuansheng Hua, and Xiao Xiang Zhu,~\IEEEmembership{Fellow,~IEEE}

\IEEEcompsocitemizethanks{

Z. Yuan and L. Mou are with the Chair of Data Science in Earth Observation, Technical University of Munich, 80333 Munich, Germany. (e-mails: zhenghang.yuan@tum.de; lichao.mou@tum.de)

Y. Hua is with the College of Civil and Transportation Engineering, Shenzhen University, Shenzhen 518060, China. (e-mail: yuansheng.hua@szu.edu.cn)

X. X. Zhu is with the Chair of Data Science in Earth Observation, Technical University of Munich, 80333 Munich, Germany, and also with the Munich Center for Machine Learning, 80333 Munich, Germany (e-mail: xiaoxiang.zhu@tum.de).
}
}

\markboth{IEEE TRANSACTIONS ON Geoscience and Remote Sensing}%
{Shell \MakeLowercase{\textit{et al.}}: Bare Demo of IEEEtran.cls for IEEE Journals}

\maketitle

\begin{abstract}
Localizing desired objects from remote sensing images is of great use in practical applications. Referring image segmentation, which aims at segmenting out the objects to which a given expression refers, has been extensively studied in natural images. However, almost no research attention is given to this task of remote sensing imagery. Considering its potential for real-world applications, in this paper, we introduce referring remote sensing image segmentation (RRSIS) to fill in this gap and make some insightful explorations. Specifically, we create a new dataset, called RefSegRS, for this task, enabling us to evaluate different methods. Afterward, we benchmark referring image segmentation methods of natural images on the RefSegRS dataset and find that these models show limited efficacy in detecting small and scattered objects. To alleviate this issue, we propose a language-guided cross-scale enhancement (LGCE) module that utilizes linguistic features to adaptively enhance multi-scale visual features by integrating both deep and shallow features. The proposed dataset, benchmarking results, and the designed LGCE module provide insights into the design of a better RRSIS model. The dataset and code will be available at {https://gitlab.lrz.de/ai4eo/reasoning/rrsis}.

\end{abstract}

\section{Introduction}
\label{sec:intro}

\IEEEPARstart{I}n recent years, the volume of remote sensing data has grown considerably. The surge in data volume has prompted the development of various techniques aiming at automatic perception of remote sensing imagery. During the early stages, traditional approaches, such as bag-of-words-based methods \cite{yang2010bag}, are widely used in tasks including image classification \cite{yang2007evaluating} and object detection \cite{sun2011automatic}. Additionally, unanimously accepted classical methods, including JSEG \cite{deng2001unsupervised}, graph cuts \cite{boykov2001fast}, watershed \cite{roerdink2000watershed}, superpixels \cite{achanta2012slic}, and EM-based algorithms \cite{belongie1998color}, are developed and actively applied to the segmentation of remote sensing images.

\begin{figure}
	\centering
	\includegraphics[width=0.45 \textwidth]{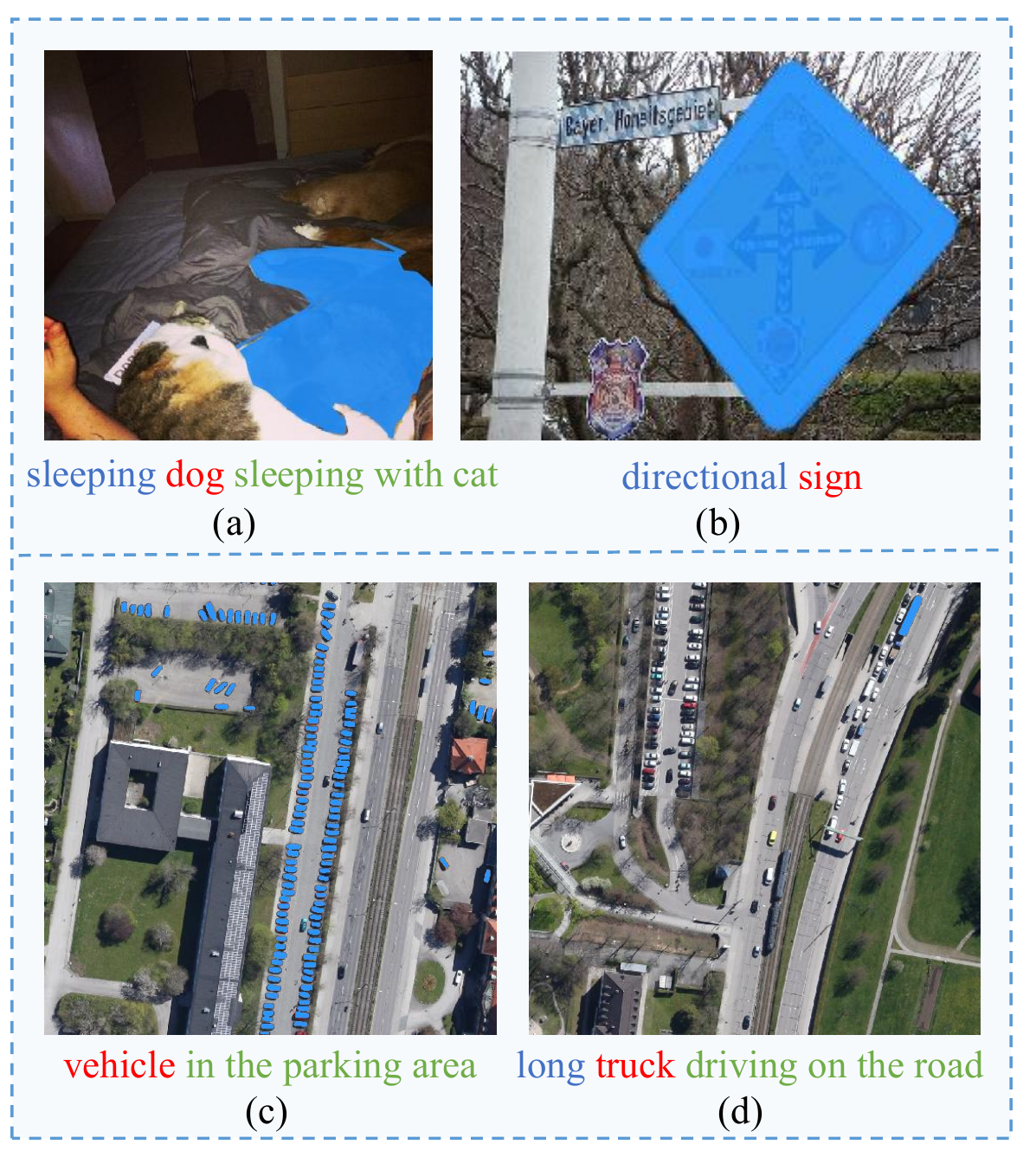}
	\caption{Examples (a) and (b) from the VGPhraseCut dataset \cite{wu2020phrasecut}, and (c) and (d) from the RefSegRS dataset. The \textcolor{red}{red}, \textcolor{blue}{blue}, and \textcolor{green}{green} highlights in referring expressions represent categories, attributes, and spatial relationships, respectively.}
	\label{FIG1}
\end{figure}

With the advent of deep learning, numerous models based on deep neural networks have been proposed, achieving dominating performance on the perception of remote sensing images. Convolutional neural networks (CNNs) such as VGG \cite{simonyan2014very}, ResNet \cite{he2016deep}, and ConvNext \cite{liu2022convnet} achieve human-competitive accuracy in image classification on large-scale datasets. More recently, Transformer-based networks like Vision Transformer (ViT) \cite{dosovitskiy2020image} and Swin Transformer \cite{liu2021swin}, introduce the capacity to model long-distance relations within images, further enhancing the image classification performance. Moreover, object detection in remote sensing images has witnessed a dramatic improvement in accuracy with the emergence of models like Faster R-CNN \cite{girshick2015fast}, YOLO \cite{redmon2016you}, and DETR \cite{carion2020end}. Simultaneously, the task of semantic segmentation has also benefited immensely from deep learning techniques, with models like FCN \cite{long2015fully}, U-Net \cite{ronneberger2015u}, Deeplab \cite{chen2017deeplab}, and Segformer \cite{xie2021segformer} attaining remarkable performance, allowing for pixel-level interpretation of remote sensing imagery.

The models tailored for the tasks mentioned above categorize elements at the levels of images, objects, or pixels, without reliance on natural language guidance. This may lead to a less targeted and suboptimal user experience. Thus, allowing end users to specify segmentation results for certain regions according to their individual needs would improve the efficiency and user interactivity of remote sensing image interpretation. In this work, we introduce referring remote sensing image segmentation (RRSIS) to provide an intuitive way for end users to specify objects of interest and enable targeted image analysis. Explicitly, given a remote sensing image and a language expression, RRSIS is to provide a pixel-level mask of desired objects based on the content of the image and the expression. Although referring image segmentation has been extensively studied in natural images \cite{ECCV16, li2018referring, CVPR19, CVPR20, CVPR22, liu2023polyformer, liu2023multi}, almost no attention has been given to this task in the context of remote sensing images. Thus, it is necessary to explore RRSIS, which enables users without expertise in the domain to obtain the precise information they require.

Derived from the concept of referring natural image segmentation, we introduce RRSIS as a novel task within the domain of remote sensing. Since there are no relevant datasets available for remote sensing data, we create a new dataset called RefSegRS on top of images and pixel-wise annotations from the SkyScapes dataset \cite{azimi2019skyscapes}. RefSegRS dataset is constructed by designing various referring expressions and generating the corresponding masks automatically. The expressions include categories, attributes, or spatial relationships with other entities, as these are the features that end-users often use to refer to objects. The dataset consists of 4,420 image-language-label triplets.
Fig. \ref{FIG1} shows two examples (natural images) from the VGPhraseCut dataset \cite{wu2020phrasecut} and two examples (remote sensing images) from the proposed dataset. Generally speaking, in natural images, targets are prominent, and scenes are small-scale, whereas this is different in remote sensing images. Due to their top-down view and limited resolution, remote sensing imagery usually contains a wide range of object categories and relatively small objects. 

In order to comprehensively explore the task of RRSIS, we evaluate a range of referring image segmentation methods \cite{ECCV16, li2018referring, CVPR19, CVPR20, CVPR22} on the RefSegRS dataset, which are originally designed for natural images.
Our experiments indicate that directly applying existing referring image segmentation models to the RefSegRS dataset does not yield satisfactory results, as we find that small and scattered objects are often missed in the results. The reason behind this is that in natural images, objects tend to be large in size, while in remote sensing images, ground objects typically occupy fewer pixels and appear small and scattered, such as cars and road markings. Therefore, to successfully identify these hard samples, it is critical to design a dedicated deep network architecture for RRSIS tasks. Transformer networks can model long-range dependencies between different parts of an image and have an advantage in detecting scattered objects (such as scattered vehicles) and identifying correlations between different object parts (such as white stripes on a zebra crossing). Thus, to address the above-mentioned issue, we base our model on language-aware vision Transformer (LAVT) \cite{CVPR22}, which is currently considered one of the state-of-the-art methods for referring image segmentation tasks.

Although LAVT can largely outperform existing CNN-based methods, experimental results show that it still struggles with segmenting small and scattered objects in remote sensing images. For this issue, in this paper, we propose a Transformer-based feature enhancement module. It is motivated by two key points regarding characteristics of deep and shallow features. On the one hand, deep features exhibit high abstraction, and each pixel has strong expressive ability, which makes it easy to categorize them into different classes. On the other hand, while shallow features have high spatial resolution, making it possible to identify small objects, their semantic level is relatively low. Taking into account the above two points, we design a module to integrate deep and shallow features by leveraging linguistic features as guidance to better segment small objects.

The main contributions of this work can be summarized as follows:

\begin{itemize}
	\item We introduce the task of RRSIS, develop a method to automatically generate ground truth masks based on natural language expressions, and create a new dataset called RefSegRS for this task. The proposed dataset consists of 4,420 image-language-label triplets, including a wide range of objects with great scale variations.
	
	\item We extensively evaluate existing referring image segmentation methods, originally designed for natural images, on the RefSegRS dataset. Based on quantitative and qualitative results, we provide a detailed analysis and valuable insights into RRSIS.
	
	\item A novel language-guided cross-scale enhancement (LGCE) module is devised to improve segmentation performance on small objects and scattered distributed objects in remote sensing images. 

\end{itemize}

The rest of the paper is organized as follows. Section \ref{Related Work} describes the related work. The methodology is introduced in Section \ref{Methodology}. Section \ref{Experiments} presents experimental results and discussion. Finally, this paper is concluded in Section \ref{Conclusion}.

\section{Related Work}
\label{Related Work}
We review the literature related to referring image segmentation on natural images as well as the combination of remote sensing imagery and natural language, including visual grounding, visual question answering (VQA), and image captioning on remote sensing data.

\subsection{Referring Image Segmentation on Natural Images}

Typically, a referring image segmentation method first extracts visual and linguistic features from images and language expressions and then fuses the multimodal features to predict masks \cite{ICCV17}. The pioneer work can be found in \cite{ECCV16}, where a long short-term memory (LSTM) \cite{LSTM1997} network is employed to encode referring expressions into vectors, and a CNN is used to learn visual features. Then a fully convolutional network (FCN) generates masks based on both modalities. This inspires subsequent works, such as using DeepLab \cite{li2018referring}, \cite{ICCV17} to learn better image representations and utilizing simple recurrent units (SRUs) \cite{margffoy2018dynamic} or bi-directional gated recurrent units (GRUs) \cite{jing2021locate} to extract better linguistic features from texts. In terms of the multimodal feature fusion, a range of attention mechanisms \cite{CVPR19}, \cite{CVPR20}, \cite{shi2018key}, \cite{feng2021encoder} are studied to model the relationship between language and image modalities. Besides, inspired by the success of contrastive
language-image pretraining (CLIP) \cite{radford2021learning}, the authors of \cite{wang2022cris} propose a CLIP-driven framework for transferring knowledge obtained from CLIP to achieve text-to-pixel alignment. 

Transformers have recently showcased significant accomplishments in computer vision tasks. 
LAVT \cite{CVPR22} employs bidirectional encoder representations from Transformers (BERT) \cite{devlin2018bert} and a novel Transformer network for referring image segmentation. BERT \cite{devlin2018bert}, a model designed for extracting language representation, is pre-trained on a considerable volume of text data. It excels at learning features from text and can be fine-tuned for various downstream tasks, showing improvements in natural language processing-related tasks. Additionally, LAVT introduces a pixel-word attention module (PWAM), which aims at aligning visual representations with language features. Specifically, given visual and language features, PWAM performs $1 \times1 $ convolutions, elements-wise multiplication, matrix multiplication, and softmax to align and fuse multi-modal features.

\subsection{Visual Grounding in Remote Sensing}
The goal of visual grounding is to localize objects mentioned in language expressions by providing bounding boxes \cite{zhan2023rsvg}. Since it is a relatively new task in remote sensing, research on it is limited. Sun et al. \cite{sun2022visual} collect a remote sensing visual grounding dataset and propose a method consisting of a language encoder, a vision encoder, and a fusion module. More specifically, the language encoder builds a relation graph for numerical geospatial relations, and the vision encoder utilizes an adaptive region attention mechanism to extract key visual information. Another work \cite{zhan2023rsvg} also creates a dataset, which is based on the existing object detection dataset DIOR \cite{li2020object}. In this work, a Transformer-based module is developed to address the challenge of background clutter and scale variation in aerial scenes.

\subsection{Visual Question Answering for Remote Sensing Data}
Given an image and a corresponding question, a VQA system aims to provide a correct answer to the question based on the image. Lobry et al. \cite{lobry2020rsvqa} first introduce VQA into remote sensing by creating two datasets for this task. They use a CNN and a recurrent neural network (RNN) for feature extraction and then fuse multi-modal features to predict answers. Subsequently, a large-scale dataset named RSVQAxBEN \cite{9553307} is established on top of the BigEarthNet dataset \cite{sumbul2019bigearthnet} to further advance the research on this task. To enhance the multi-modal representation, many methods have been investigated. The work \cite{zheng2021mutual} employs an attention mechanism and a bilinear technique to improve the alignment between visual and textual modalities. The paper \cite{9832935} designs a bi-modal Transformer-based method to learn powerful multi-modal features. In \cite{10018408}, a spatial hierarchical reasoning network is designed toward a better fusion of visual and language features. Chappuis et al. \cite{chappuis2022prompt} propose to extract context information from the visual input and inject it into a language model to obtain answers. Their method improves the performance as well as the interpretability of VQA systems. In addition, the work \cite{yuan2022change} introduces a change detection-based VQA model to extend this task to a multi-temporal domain.

\subsection{Remote Sensing Image Captioning}
Image captioning aims to describe the given images. \cite{qu2016deep} first introduces this task in remote sensing and utilizes a CNN and an RNN to extract visual features and generate words, respectively. Subsequently, \cite{lu2017exploring} provides a comprehensive review of using handcrafted and deep features in this task. Encoder-decoder architectures are used in multiple works \cite{li2020truncation, zhao2021high, li2021recurrent, zhang2019lam}, where an encoder learns visual features from the input images and a decoder transfers the features to textual descriptions. In addition, various attention-based methods are studied a lot in recent years. For example, Zhao et al. \cite{zhao2021high} develop a structured attention-based method, in order to use structural features in semantic information. Zhang et al. \cite{zhang2023multi} propose a multi-source interactive attention network to investigate the impact of the semantic attribute and contextual information of generated words in producing descriptions. More recently, Transformers are also explored in this task. \cite{9855519} exploits a Transformer network as an encoder-decoder and an auxiliary decoder to address the limited data issue. A multilayer aggregated Transformer is designed in \cite{9709791} to leverage multi-scale features in the decoder.

\begin{figure*}
	\centering
	\includegraphics[width=0.9 \textwidth]{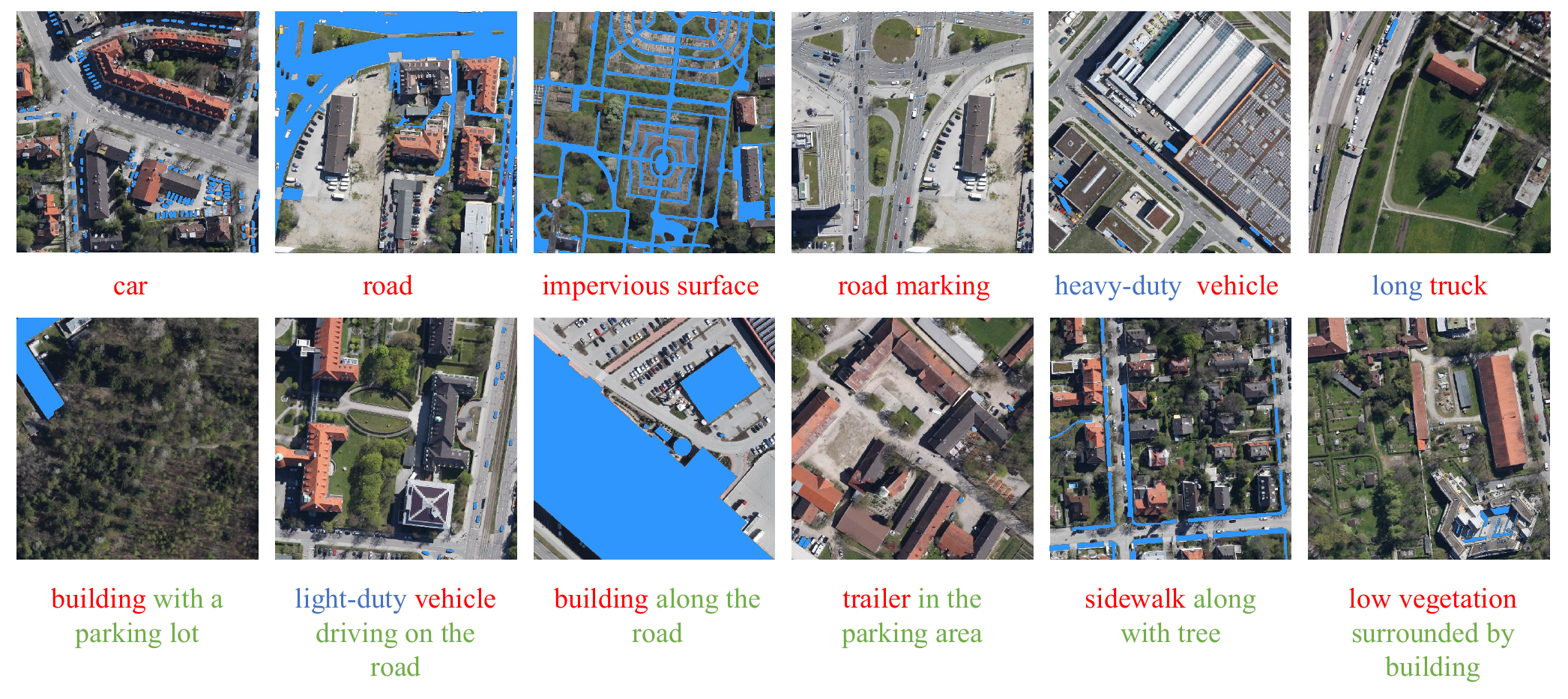}
	\caption{Visualization examples of the proposed dataset. For a distinct visualization, the corresponding masks are superimposed on the original images. The \textcolor{red}{red}, \textcolor{blue}{blue}, and \textcolor{green}{green} highlights in referring expressions represent categories, attributes, and spatial relationships, respectively.}
	\label{FIG3}
\end{figure*}

\section{Dataset Construction}
\label{Dataset}
In this section, we outline the process of collecting images and generating referring expressions for the construction of the  RefSegRS dataset. 

\subsection{Image Collection}
We collect images by cropping large tiles from the SkyScapes dataset \cite{azimi2019skyscapes}, which consists of 16 RGB tiles. Each tile has a size of $5616 \times 3744$ pixels with a spatial resolution of 13 cm. All pixels are divided into 20 categories: low vegetation, paved road, non-paved road, paved parking place, non-paved parking place, bikeway, sidewalk, entrance/exit, danger area, lane marking, building, car, trailer, van, truck, large truck, bus, clutter, impervious surface, and tree. For those labeled as lane marking, the creators also offer their specific types such as dash-line, long-line, and small dash-line, but these labels are not considered in our case. To ensure that each image contains adequate objects and inter-object relations, we crop the tiles into images of $1200 \times 1200$ pixels using a sliding window with a stride of 600 pixels. Considering the input size of deep neural networks, we further downsample them into $512 \times 512$ pixels. Fig.~\ref{FIG3} shows several example images from the RefSegRS dataset.

\begin{table}
	\centering
	\caption{Features of the RefSegRS Dataset.}
	\scalebox{1}{
		\begin{tabular}{m{2cm}<{} m{5cm}<{} }
			\toprule
			Feature           & Label          \\  \midrule
			category            & road, vehicle, car, van, building, truck, trailer, bus, road marking, bikeway, sidewalk, tree, low vegetation, impervious surface  \\  \midrule
   attribute            & paved, unpaved, light-duty, heavy-duty, long     \\  \midrule
		  spatial relation             & in the parking area, with a parking lot, driving on the road, along the road, along with tree, on the road, surrounded by building \\ 	
			\bottomrule
		\end{tabular}
	}\label{tabel-1}
\end{table}

\subsection{Referring Expression Generation}
Given that end users often refer to objects by their categories, attributes, and spatial relationships with other entities, we generate linguistic expressions using the following templates.

\begin{itemize}
\item \textit{Category or category with attribute}. This is a commonly used expression for common users to indicate their interests in certain objects. Based on this template, we generate referring expressions by 1) directly specifying categories such as ``vehicle'' and ``road” or 2) adding attributes to them, e.g., ``light-weight vehicle''. After analyzing the SkyScapes dataset, we identify 14 distinct categories and 5 attribute tags that can be used to describe objects, as shown in Table \ref{tabel-1}.	
\item \textit{Category with spatial relation}. Spatial relationships among objects are another feature that end users often want to explore. Given that remote sensing images are taken from a top-down view, we primarily focus on two spatial relationships: adjacency and containment. For the former, we make use of words such as ``with", ``along", and ``along with". For instance, a generated expression could be ``building along the road". For the latter, we use phrases like ``surrounded by", ``in", and ``on". An example here is ``low vegetation surrounded by building". There are a total of seven different spatial relationships that we consider in our dataset.
\item \textit{Category with attribute and spatial relation}. This template includes both attribute-related and spatial relation-related descriptions. For example, ``light-duty vehicle in the parking area". 
\end{itemize}

\begin{figure}
	\centering
	\includegraphics[width=0.45 \textwidth]{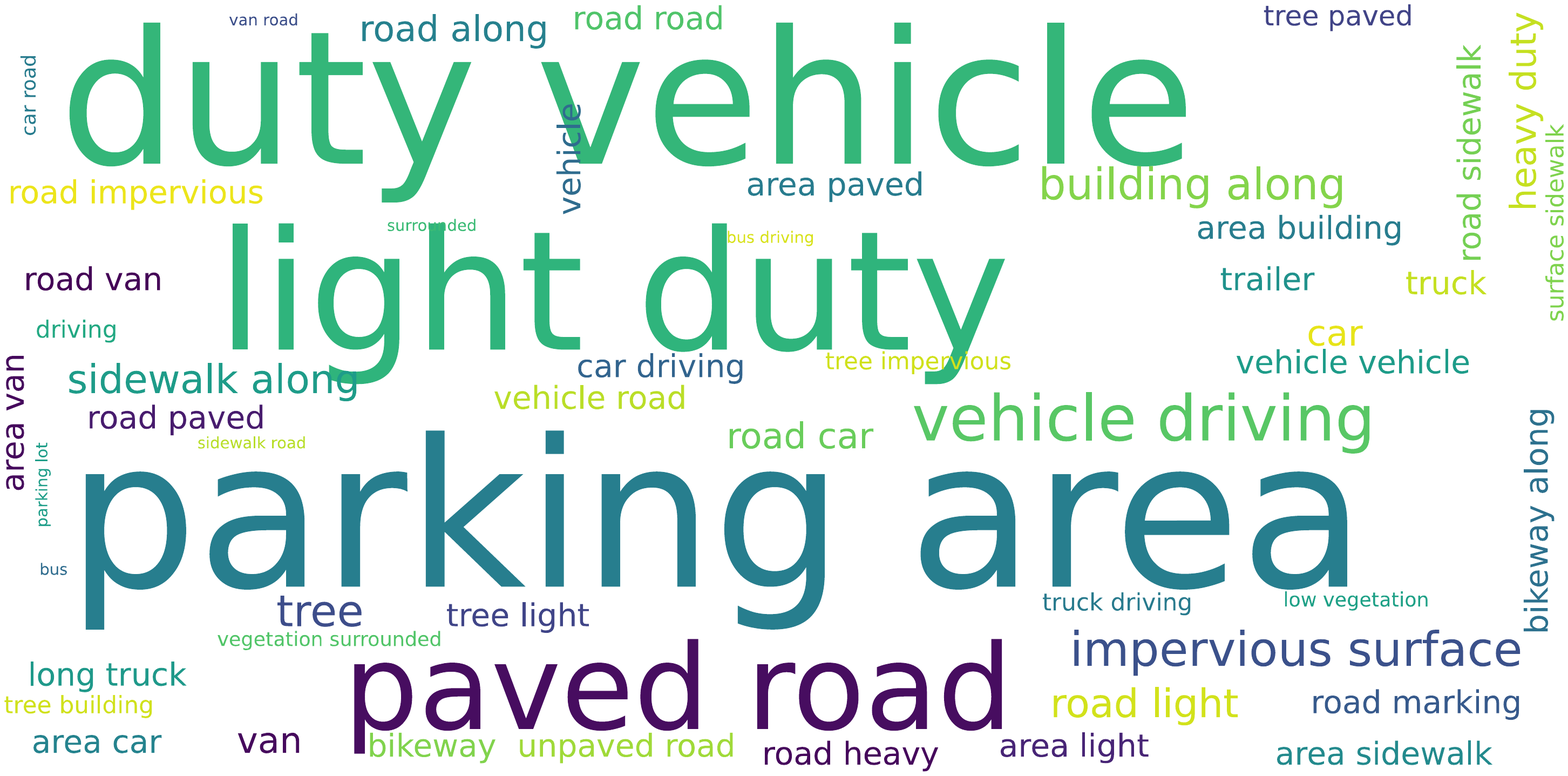}
	\caption{Word cloud for referring expressions in the RefSegRS dataset.}
	\label{FIG2}
\end{figure}

To offer a comprehensive visualization of the primary expressions within the dataset, we present a word cloud of the generated expressions in Fig. \ref{FIG2}. As shown in it, “parking area” emerges as the most frequently recurring object in the dataset.

\begin{figure*}
	\centering
	\includegraphics[width=0.95 \textwidth]{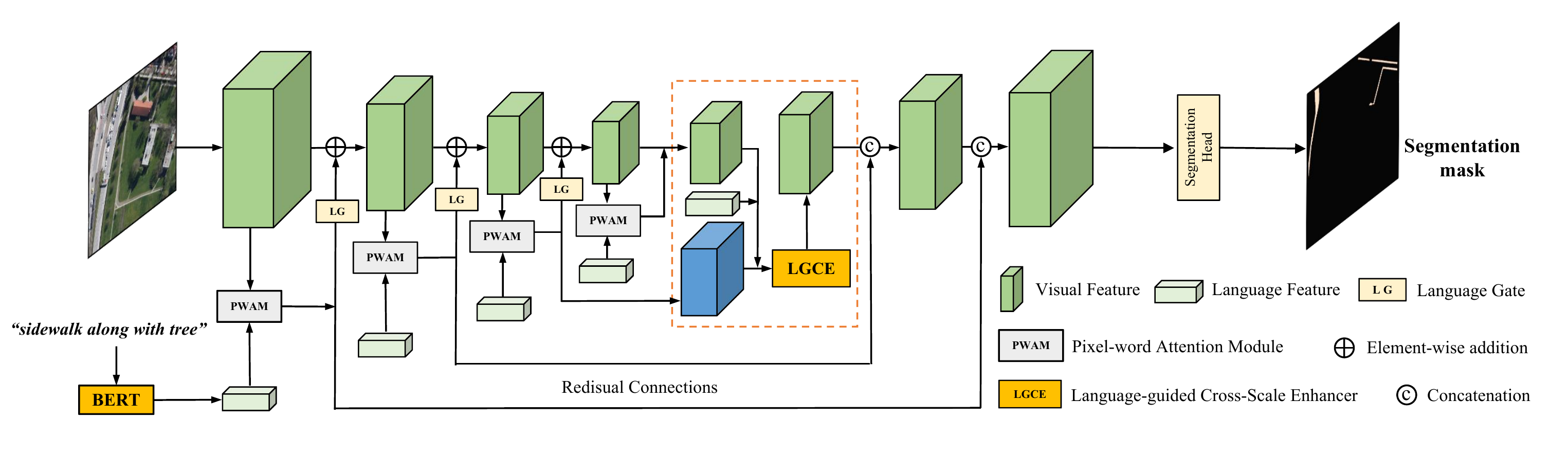}
	\caption{Overall architecture of our RRSIS model.} 
	\label{framework}
\end{figure*}

\subsection{Mask Generation}
Segmentation maps are sourced from the SkyScapes dataset \cite{azimi2019skyscapes}, with each pixel in the maps indicating a class label. They are used to automatically generate ground truth masks (binary masks) based on natural language expressions for referring image segmentation tasks.
In order to generate pixel-wise annotations for language expressions, we take the following steps:

\begin{itemize}
    \item \textit{Step 1.} We conclude two types of conceptual relationships between the generated expressions and the original labels in the Skyscapes dataset: identity and inclusion. Here are two examples: ``road marking'' is identical to ``lane marking" but more idiomatic, while ``light-duty vehicle'' is the collection of ``car" and ``van".
    
    \item \textit{Step 2.} To generate a binary mask for each category, we make use of pixel-wise annotations in the SkyScapes dataset and set values of pixels within/outside the category to 1/0. For categories composed of several subcategories (e.g., “vehicle” includes “car”, “van”, “truck”, etc.), we perform element-wise additions on binary masks of subcategories automatically.
    
    \item \textit{Step 3.} For expressions generated using spatial relationships, we partition masks obtained from step 2 into multiple instances (i.e., connected regions) and subsequently remove instances that do not conform to the specified spatial relationships. We also add buffers to these instances in determining whether they are adjacent or not, so that outputs are more in line with human intuition visually. Afterward, the remaining instances are gathered for producing the final masks of expressions.
\end{itemize}

\subsection{Dataset Statistics}

We conduct spatial relation and attribute analysis on semantic labels of the SkyScapes dataset to generate natural language expressions and the corresponding masks. After manually filtering out uninformative image-language-label triplets, we ultimately obtain a dataset consisting of 4,420 image-language-label triplets across 285 scenes. There are 151 scenes with 2,172 referring expressions in the training set, 31 scenes with 431 expressions in the validation set, and 103 scenes with 1,817 expressions in the test set. Note that referring expressions associated with the same scene are grouped into one set, which means that there is no overlap among the three subsets.

\section{Methodology}
\label{Methodology}
Referring image segmentation aims at segmenting out target objects with the guidance of natural language. To this end, representation learning, multi-modal feature fusion, and semantic segmentation are crucial components of this system. A typical referring image segmentation model usually consists of four parts: a visual encoder, a language encoder, a multi-modal feature fusion module, and a segmentation head \cite{ECCV16}. Specifically, in the stage of feature learning, visual features of a given image are extracted by the visual encoder, and linguistic features of the input text are computed by the language encoder. Subsequently, the multi-modal fusion module is used to integrate information from both vision and language modalities via a sophisticated fusion mechanism. In the end, the multi-modal features are fed into the segmentation head to predict the final mask. By doing so, the corresponding objects described in the referring expression can be outlined by the model. 

Thanks to the long-distance spatial modeling and multi-modal learning abilities of Transformers \cite{vaswani2017attention, dosovitskiy2020image}, the performance of pixel-level dense prediction tasks has been greatly enhanced in both computer vision and remote sensing communities. Thus, we base our baseline method on a Transformer-based model LAVT \cite{CVPR22}. Further, we design an LGCE module to improve the identification of small objects and scattered distributed objects in remote
sensing images.

 \begin{figure*}
	\centering
	\includegraphics[width=0.92 \textwidth]{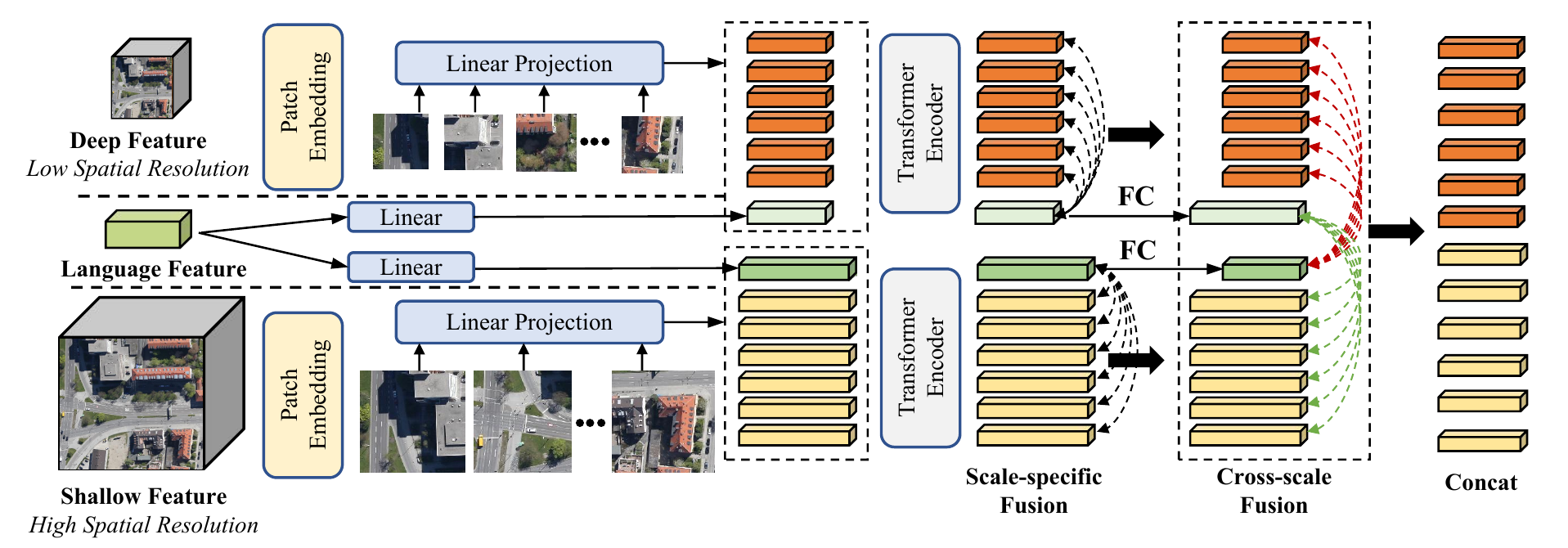}
	\caption{The proposed LGCE module. It aims at effectively integrating deep and shallow features by leveraging language guidance.}
	\label{module}
\end{figure*}

In the following subsections, we start with the introduction of the whole framework of the used RRSIS model (see Fig. \ref{framework}). Then we give a detailed description of the proposed LGCE module.

\subsection{Overall Architecture}
We adopt LAVT as our baseline framework. Following it, we use Swin Transformer \cite{liu2021swin} as the backbone network for extracting visual features from the input image. 
Since language has a guiding function in achieving final results, it is necessary to learn semantic representations of language expressions. Hence, we incorporate BERT \cite{devlin2018bert} to extract representative language embeddings. BERT is trained using two methods: 1) masked language modeling, predicting randomly masked words in a sentence; 2) next sentence prediction, determining if two sentences logically follow each other. This dual-training approach empowers BERT to understand complex languages, making it effective for diverse natural language processing tasks. In RRSIS tasks, BERT is employed for three advantages. First, BERT can help handle ambiguity and variability in natural language inputs. The representative language encoding enables the model to capture and understand the semantic meaning in the given language guidance. Second, the utilization of language embeddings from BERT offers scalability and flexibility. It allows the model to understand and process a wide range of language inputs and generalize to unseen input sentences. Third, the pre-trained language embeddings from BERT can enhance the efficiency of model training and improve performance. Regarding multi-modal fusion, LAVT designs a PWAM and a language gate (LG) module to fuse language and visual features.

For the part of feature decoding, LAVT exploits a UNet-like architecture to enhance feature maps in the decoding stage with the encoded features by using simple residual connections. Although it has been proven to be effective for natural images, we argue that this simple combination of features with different spatial sizes is not optimal for remote sensing images. We observe that expressions accompanying such images often contain information about the size of target objects. For example, texts containing car and road marking tend to refer to small objects, while texts containing building generally relate to large objects. The accurate identification of small objects requires a more sophisticated multi-scale feature fusion module that takes the input text expressions into account. To this end, we design a language-guided Transformer module that enhances visual representations. As the proposed method is built upon a Transformer architecture, to clearly explain our approach, we first briefly describe ViT \cite{dosovitskiy2020image} in the following subsection.

\subsection{Vision Transformer}
Different from CNNs, ViT first converts image patches into tokens by a linear patch embedding layer. Considering that the computation of self-attention is position-agnostic, ViT adds additional position embeddings to the patch tokens. Afterward, the tokens are fed into Transformer encoders for further processing. For each block in a Transformer encoder, the multi-head self-attention and layer normalization are first used to transform the tokens. Then, a multi-layer perceptron with the Gaussian error linear unit (GELU) non-linearity \cite{hendrycks2016gaussian} is employed to generate the output of the Transformer block. Formally, let $\bm{x}$ denote visual tokens and $\bm{{E}}_{pos}$ represent position embeddings. The computation process of the input tokens and the ${\ell}$-th Transformer block can be described as:
\begin{equation}
\begin{split}
\bm{z}_0 &= \text{PE}(\bm{x}) + \bm{{E}}_{pos}, \\
{\bm{{z}_{\ell}}\prime} &= \bm{z}_{\ell-1} + \MSA(\LN(\bm{z}_{\ell-1})), \, \, \ell=1\dotsb L,  \\
\bm{z}_{\ell} &= {\bm{z}_{\ell}}\prime + \MLP(\LN({\bm{z}_{\ell}}\prime)), \, \, \ell=1\dotsb L,\\
\end{split}
\label{TL}
\end{equation}
where $\bm{z}_0$ is the input to the Transformer encoder. PE$(\cdot)$, MSA$(\cdot)$, LN$(\cdot)$, and MLP$(\cdot)$ stand for the patch embedding layer, multi-head self-attention, layer normalization, and multi-layer perceptron, respectively. ${\bm{{z}_{\ell}}\prime}$ and $\bm{z}_{\ell}$ represent outputs of MSA and MLP from the $\ell$-th Transformer block.  Note that residual connections are used in Transformer blocks.

\begin{table*}[]
\centering
	 \caption{Numerical Results of Comparisons with Existing Methods on the RefSegRS Dataset. The best Performance is Bold.}
 \resizebox{15cm}{!}{
\begin{tabular}{ccccccccc}
\toprule
Methods  & Publication & Pr@0.5          & Pr@0.6          & Pr@0.7          & Pr@0.8          & Pr@0.9         & oIoU            & mIoU            \\  \midrule
    LSTM-CNN \cite{ECCV16} & ECCV'16      & 0.1569          & 0.1057          & 0.0517          & 0.0110          & 0.0028         & 0.5383          & 0.2476          \\ 
ConvLSTM \cite{li2018referring} & CVPR'18      & 0.3121          & 0.2339          & 0.1530          & 0.0759          & 0.0110         & 0.6612          & 0.4334          \\ 
CMSA  \cite{CVPR19}    & CVPR'19      & 0.2807          & 0.2025          & 0.1271          & 0.0561          & 0.0083         & 0.6453          & 0.4147          \\ 
BRINet \cite{CVPR20}  & CVPR'20      & 0.2256          & 0.1574          & 0.0985          & 0.0352          & 0.0050         & 0.6016          & 0.3287          \\ 
LAVT   \cite{CVPR22}     & CVPR'22      & 0.7144          & 0.5740          & 0.3214          & 0.1541          & 0.0451          & 0.7646          & 0.5774          \\ 
Ours     & ---            & \textbf{0.7375} & \textbf{0.6114} & \textbf{0.3946} & \textbf{0.1602} & \textbf{0.0545} & \textbf{0.7681} & \textbf{0.5996} \\  \bottomrule
\end{tabular}
} \label{exp_1}
\end{table*}

\subsection{Language-guided Cross-scale Enhancement Module}
The main motivation for designing the LGCE module is to leverage language as guidance to adaptively enhance multi-scale visual features. To formally describe LGCE, we denote the encoded visual features from four stages of Swin Transformer as $\bm{V}^i \in \mathbb{R}^{C_i\times H_i\times W_i}, i \in \{1,2,3,4\}$, where $C_i, H_i$, and $W_i$ represent the number of channels, height, and width of feature maps from the $i$-th stage. Among them, $\bm{V}^4$ is the output of the fourth stage with the highest abstraction but the lowest spatial resolution. Language feature vectors encoded by BERT \cite{devlin2018bert} can be denoted as $\bm{L} \in \mathbb{R}^{C_t \times T}$, where $C_t$ and $T$ represent the number of channels and the count of words, respectively.

The LGCE module takes $\bm{V}^3$ and $\bm{V}^4$ as the input features with different spatial resolutions. $\bm{V}^3$ represents shallow features with a high spatial resolution, and $\bm{V}^4$ denotes deep features with a relatively low spatial resolution.
In order to efficiently fuse language and visual features, we first average language features ${\bm{L} \in \mathbb{R}^{C_t \times T}}$ along the last dimension to obtain the mean language feature $\Bar{\bm{L}} \in \mathbb{R}^{C_t}$. Subsequently, two separate linear projections $f_{h}$ and $f_{l}$ are performed on $\Bar{\bm{L}}$ to match dimensions of $\bm{V}^3$ and $\bm{V}^4$. Then, $f_{h}(\Bar{\bm{L}})$ and $f_{l}(\Bar{\bm{L}})$ are concatenated with $\bm{V}^3$ and $\bm{V}^4$, respectively, which is shown by the scale-specific fusion part in Fig. \ref{module}. The calculation process can be written as
\begin{equation}
\begin{split}
    \bm{z}_{h} &= \text{TL}_{h}(f_{h}(\Bar{\bm{L}}) \parallel \bm{V}^3),\\
    \bm{z}_l &= \text{TL}_l(f_{l}(\Bar{\bm{L}}) \parallel \bm{V}^4),
\end{split}
\end{equation}
where $\parallel$ denotes concatenation.
$\text{TL}_h$ and $\text{TL}_l$ are Transformer blocks. 
$\bm{z}_h$ and $\bm{z}_l$ are output features with high and low resolutions, respectively, and both are feature vectors that contain language and visual features (shown in green and yellow blocks in Fig. \ref{module}). 

Then, the split operation is applied to $\bm{z}_h$ and $\bm{z}_l$, dividing them into two feature components. This operation serves as the inverse of the concatenation operation, which can be defined as:
\begin{equation}
\begin{split}
&\bm{L}_h, \bm{V}_h = \text{split}(\bm{z}_h), \\
&\bm{L}_l, \bm{V}_l = \text{split}(\bm{z}_l), \\
\end{split}
\end{equation}
where $\text{split}(\cdot)$ is utilized to divide the given data into two subsets.
$\bm{L}_h$ and $\bm{L}_l$ denote output language features of $\text{TL}_h$ and $\text{TL}_l$, respectively. $\bm{V}_h$ and $\bm{V}_l$ represent output visual features.
Next, we use $\bm{L}_h$ and $\bm{L}_l$ as intermediary tokens to perform cross-scale fusion. Two fully connected (FC) layers ${f_h}\prime$ and ${f_l}\prime$ are employed to align dimensions of language features with those of visual features. Specifically, $\bm{L}_h$ is aligned with $\bm{V}_l$, and $\bm{L}_l$ is aligned with $\bm{V}_h$.
Further, two multi-head self-attention layers $\text{MSA}_h$ and $\text{MSA}_l$ are exploited to fuse features with different spatial resolutions in Transformer layers. The cross-scale fusion can be formulated as
\begin{equation}
\begin{split}
&{\bm{z}_h}\prime = {f_l}\prime(\bm{L}_l) + \text{MSA}_h(\text{LN}({f_l}\prime(\bm{L}_l) \parallel \bm{V}_h)),\\
&{\bm{z}_l}\prime = {f_h}\prime(\bm{L}_h) + \text{MSA}_l(\text{LN}({f_h}\prime(\bm{L}_h) \parallel \bm{V}_l)),
\end{split}
\end{equation}
where $\bm{z}_h\prime$ and $\bm{z}_l\prime$ are output features of the cross-scale fusion. Similarly, we split $\bm{z}_h\prime$ and $\bm{z}_l\prime$ to obtain the final output features ${\bm{V}_h}\prime$ and ${\bm{V}_l}\prime$, which can be described as:
\begin{equation}
\begin{split}
& {\bm{L}_h}\prime, {\bm{V}_h}\prime=\text{split}({\bm{z}_h}\prime), \\
&{\bm{L}_l}\prime, {\bm{V}_l}\prime=\text{split}({\bm{z}_l}\prime). \\
\end{split}
\end{equation}

As shown in Fig. \ref{module}, ${\bm{V}_h}\prime$ and ${\bm{V}_l}\prime$ are concatenated and then fed into the subsequent decoding layers. Afterward, we employ a segmentation head to predict masks. Following LAVT, the feature decoding part exploits the previous output from the PWAM module in a recursive manner. Two $3\times 3$ convolutional layers with batch normalization \cite{ioffe2015batch} and ReLU are used to construct the feature decoder. Ultimately, a binary mask is predicted by a $1\times 1$ convolution layer.

\begin{table*}[]
\centering
	 \caption{Numerical Results of Ablation Experiments on the RefSegRS Dataset. The Best Performance is Bold.}
 
 \resizebox{14cm}{!}{
\begin{tabular}{cccccccc}
\toprule
Methods & Pr@0.5          & Pr@0.6          & Pr@0.7          & Pr@0.8          & Pr@0.9         & oIoU            & mIoU            \\  \midrule
     Baseline  (LAVT)     & 0.7144          & 0.5740          & 0.3214          & 0.1541          & 0.0451          & 0.7646          & 0.5774      \\  

    w/o Language Guidance     & 0.5949          & 0.3869          & 0.2042         & 0.1040         & 0.0237         & 0.7158          & 0.5140         \\ 

   w/o Cross-scale Fusion       & 0.7335          & 0.5977         & 0.3643          & 0.1579          & 0.0413         &0.7616          & 0.5856    \\    

Full Model (LGCE)            & \textbf{0.7375} & \textbf{0.6114} & \textbf{0.3946} & \textbf{0.1602} & \textbf{0.0545} & \textbf{0.7681} & \textbf{0.5996} 
\\  \bottomrule
\end{tabular}
}\label{exp_2}
\end{table*}

\begin{table}[]
\centering
	 \caption{Numerical Results of Computational Complexity on the RefSegRS Dataset. }
 
 \resizebox{8cm}{!}{
\begin{tabular}{cccc}
\toprule
Methods                                & FLOPs          &PARAMs          & MACs           \\  \midrule
Baseline (LAVT)                                & 384.27G          & 118.85M          & 191.91G                 \\  

w/o Language Guidance     & 401.34G          & 167.37M          & 200.44G              \\ 

w/o Cross-scale Fusion       & 399.44G          &160.48M        & 199.49G                \\    

Full Model (LGCE)            & 401.34G     & 167.37M    & 200.44G \\  \bottomrule
\end{tabular}
}\label{exp_3}
\end{table}

\section{Experiments}
\label{Experiments}

\subsection{Dataset and Evaluation Metrics}
             
The RefSegRS dataset consists of 4,420 image-language-label triplets, with 2172 triplets in the training set, 431 triplets in the validation set, and 1817 triplets in the test set. 
The commonly used metrics in referring image segmentation tasks include overall intersection-over-union (oIoU), mean intersection-over-union (mIoU), and precision at threshold values from 0.5 to 0.9. oIoU is calculated by taking the ratio of the total intersection area to the total union area across all test samples. mIoU is obtained by averaging IoU values between predicted masks and ground truths across all test samples. 
oIoU tends to give higher importance to larger objects while mIoU considers small and large objects equivalently.
Precision at different threshold values measures the proportion of test samples that satisfy a specified IoU threshold.

\subsection{Competing Methods}
Currently, there are no referring image segmentation approaches tailored for remote sensing data.
To explore this task, we evaluate five existing methods, originally designed for natural images, including LSTM-CNN \cite{ECCV16}, ConvLSTM \cite{li2018referring}, CMSA \cite{CVPR19}, BRINet \cite{CVPR20}, and LAVT \cite{CVPR22}, on the proposed dataset. Below, we provide a brief introduction to these methods.

\begin{itemize}
    \item LSTM-CNN \cite{ECCV16}: It consists of an LSTM \cite{LSTM1997} network for natural language encoding, the VGG-16 \cite{simonyan2014very} for visual feature extraction, and a fully convolutional upsampling module for pixel-wise segmentation. 
    
    \item ConvLSTM \cite{li2018referring}: It adopts the ResNet-101 \cite{he2016deep} to learn visual features and an LSTM to encode referring expressions. Subsequently, spatial coordinates, visual features, and language features are concatenated together as multi-modal features. Then, multiple convolutional LSTM layers are utilized to refine multi-modal features progressively.
    
    \item CMSA \cite{CVPR19}: It utilizes the ResNet-101 for visual feature encoding and keeps individual word vectors. Next, a cross-modal self-attention module is proposed to effectively capture long-range dependencies between words and spatial regions.
    
    \item BRINet \cite{CVPR20}: It employs the ResNet-101 to encode visual features and an LSTM to extract linguistic features. Then, a bi-directional cross-modal attention module is utilized to learn cross-modal relationships between linguistic and visual features. 
    
    \item LAVT \cite{CVPR22}: It uses the Swin Transformer \cite{liu2021swin} to extract visual features. For the encoding of linguistic features, LAVT uses ${BERT}_{BASE}$ \cite{devlin2018bert} as the language encoder. As for the fusion of multi-modal features, LAVT integrates the input referring expression into visual features at intermediate levels in a vision Transformer network.
\end{itemize}

\subsection{Implementation Details}

CNN-based methods, including LSTM-CNN \cite{ECCV16}, ConvLSTM\cite{li2018referring}, CMSA\cite{CVPR19}, and BRINet \cite{CVPR20}, are implemented using TensorFlow. Adam optimizer is used with an initial learning rate of 0.00025 for model training. The models are trained for 400,000 iterations for all experiments, and the batch size is set to 1. To enable a fair comparison, the hyperparameters employed for model training are maintained consistently across different models.

Transformer-based models, namely LAVT \cite{CVPR22} and the proposed method, are implemented using Pytorch. We set the batch size to 8 at training. The AdamW optimizer \cite{loshchilov2018decoupled} is used, and the initial learning rate is set to 0.00005. In addition, a weight decay of 0.01 is used for model training, and a window size of 12 is set for the Swin Transformer. For both methods, we train them for 40 epochs and select the best-performing checkpoint according to the performance on the validation set. 

\subsection{Comparisons with Existing Methods}

\begin{figure*}
	\centering
	\includegraphics[width=0.90 \textwidth]{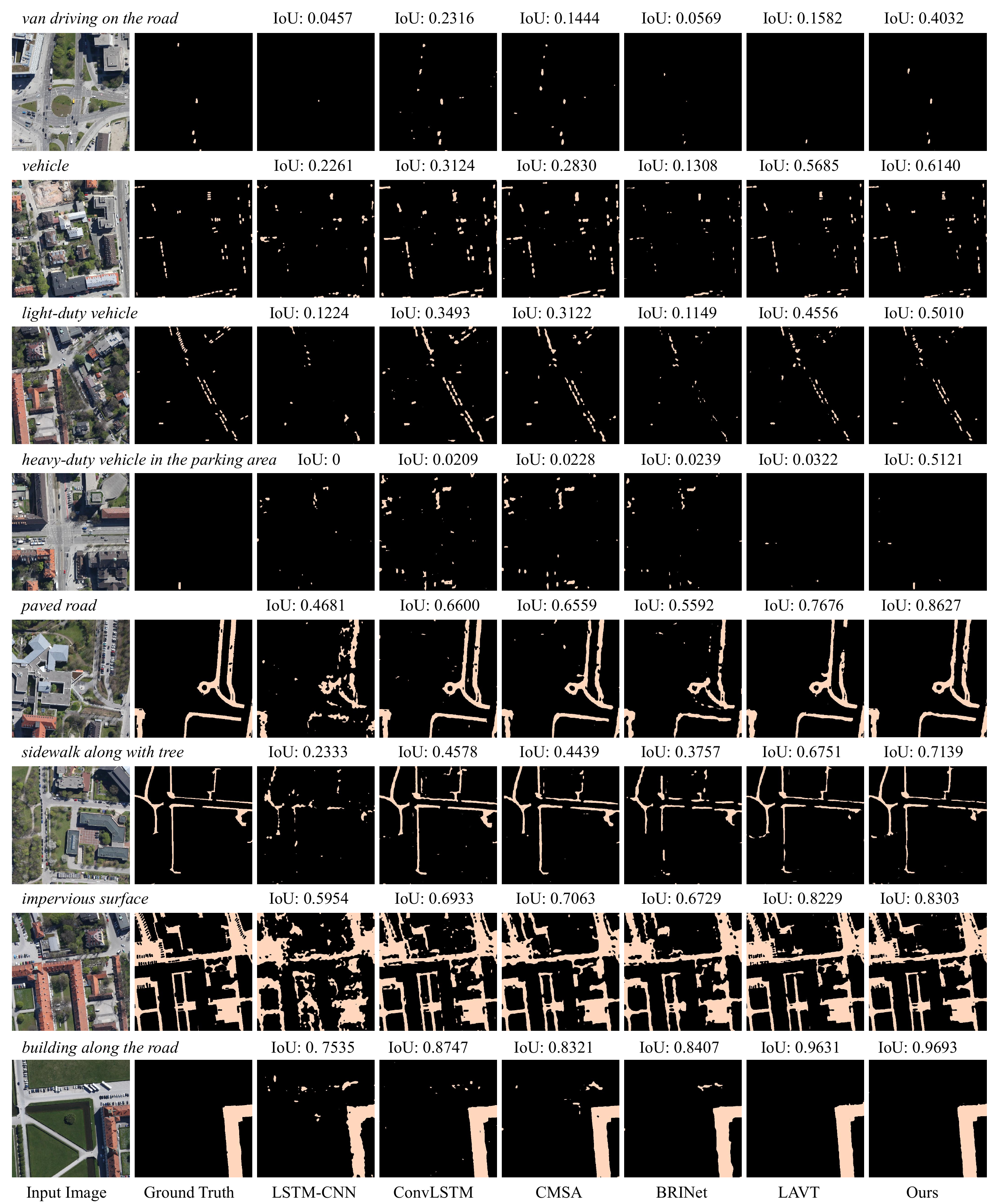}
	\caption{Illustration of some segmentation results and their corresponding IoU scores from different methods. From left to right: input image, ground truth, LSTM-CNN, ConvLSTM, CMSA, BRINet, LAVT, and our model.}
	\label{visual0}
\end{figure*}

\begin{figure}
	\centering
	\includegraphics[width=0.45 \textwidth]{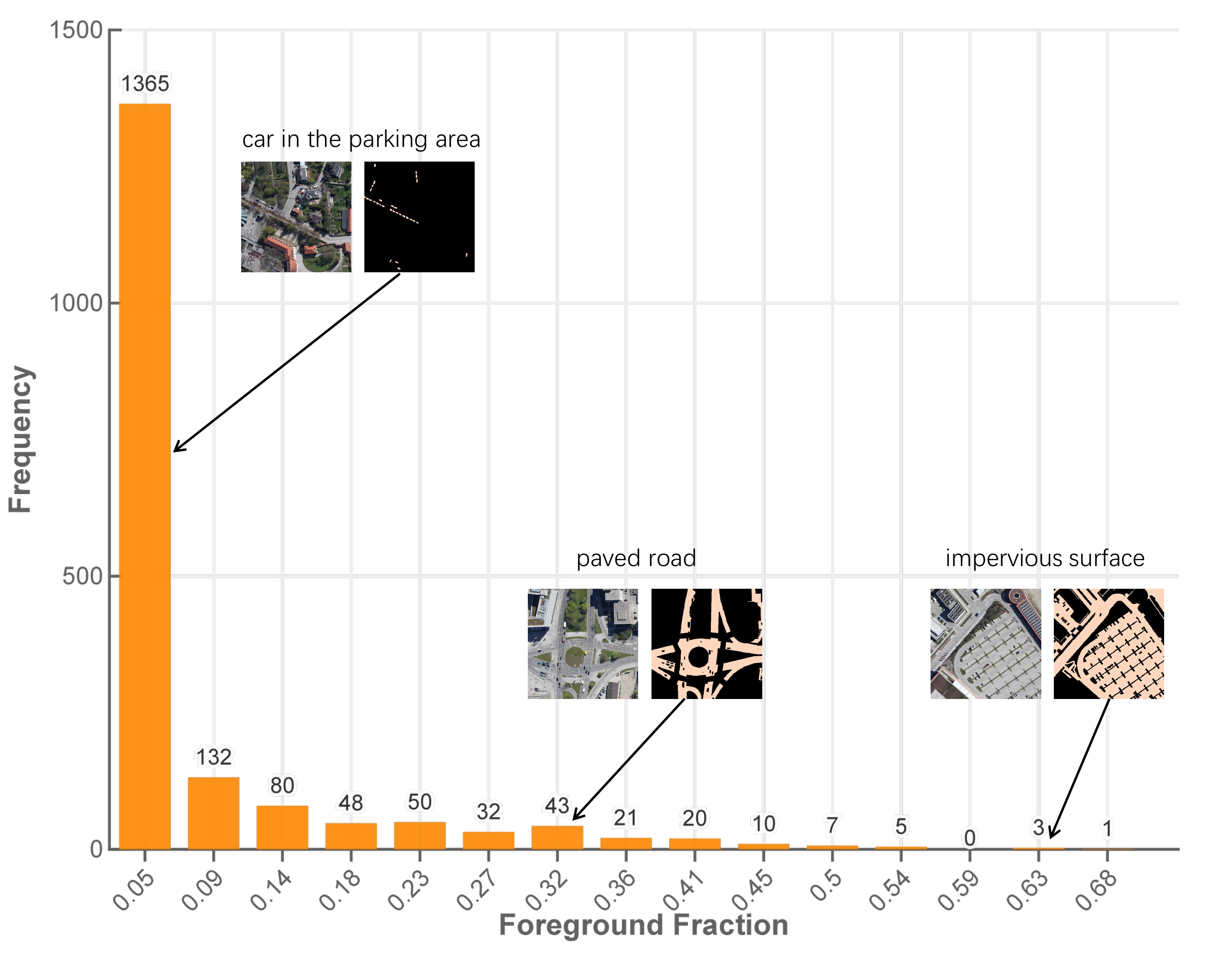}
	\caption{Histogram distribution of foreground proportions of samples in the RefSegRS dataset. Three examples with varying foreground proportions are visualized.}
	\label{h_1} 
\end{figure}

\begin{figure*}
	\centering
	\includegraphics[width=0.90 \textwidth]{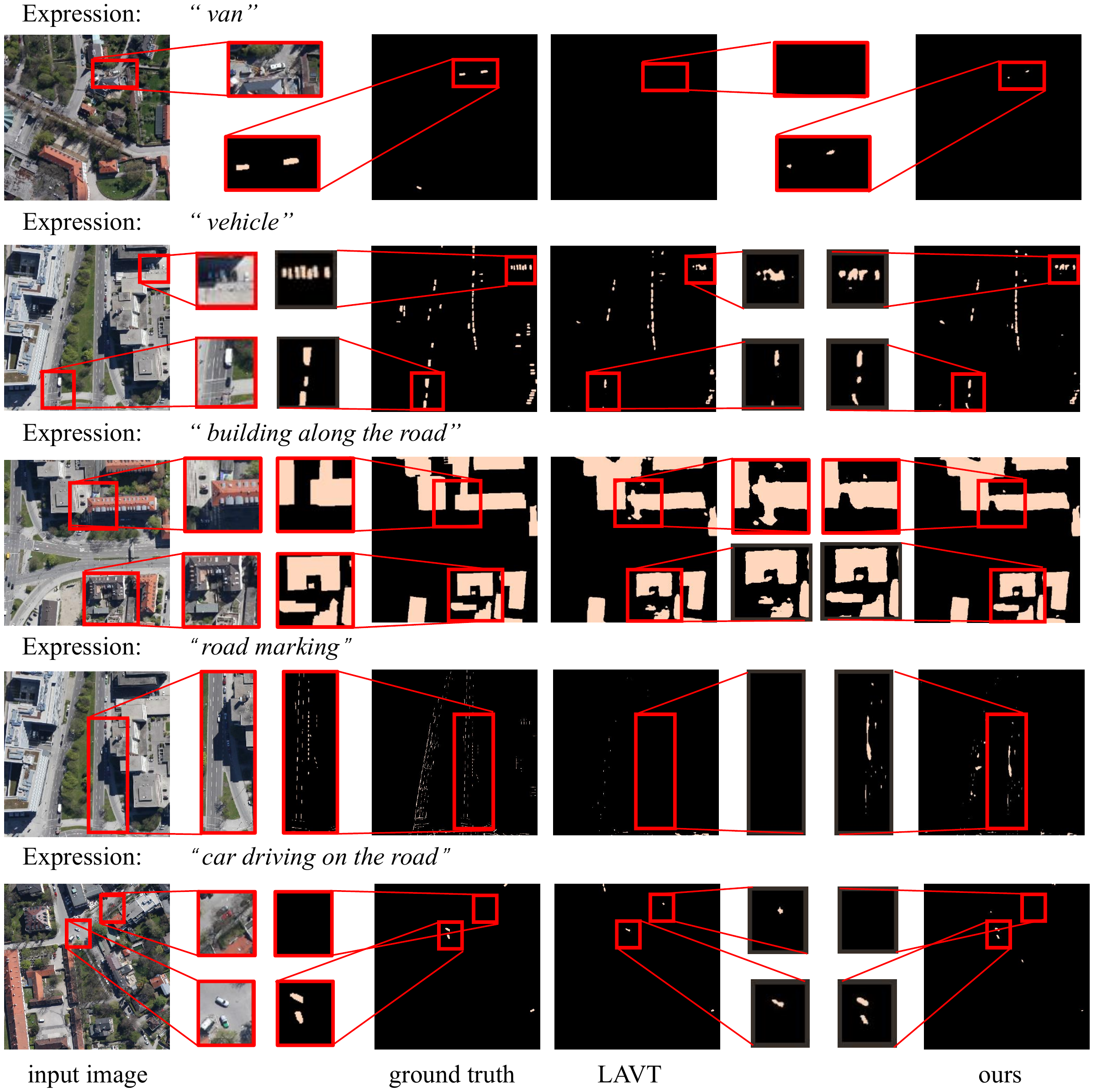}
	\caption{Examples of predictions generated by LAVT and our method on the RefSegRS dataset. Some details are outlined in red boxes for the clarity of comparison.}
	\label{visual1} 
\end{figure*}

To show the effectiveness of our method, we compare the proposed method with the aforementioned five existing methods. The results are presented in Table \ref{exp_1}.
\subsubsection{Comparison with CNN-based methods}
In general, Transformer-based methods greatly outperform CNN-based methods with regard to metrics Pr@0.5 and Pr@0.6. This indicates that there exist many hard examples, on which IoU scores of CNN-based methods are low. For instance, ConvLSTM obtains a Pr@0.5 score of 31.21\%, which means that only 31.21\% of test samples have an IoU greater than 0.5. Our Transformer-based method can achieve a Pr@0.5 of 73.75\%, which demonstrates a significant performance gap between them.

To clearly demonstrate the effectiveness of the proposed method, we opt to visually compare segmentation results and their corresponding IoU scores generated by different methods. As shown in Fig. \ref{visual0}, targets in vehicle-related samples are extremely cluttered and unevenly distributed. For these examples, CNNs are unable to distinguish small objects due to: 1) the lack of spatial details in features of deep CNN layers; 2) their weak relationship modeling ability. Although multi-level visual features are used in CNN-based methods, their performance on hard examples is still not satisfactory. 

To provide a clear overview of the test set, we count the proportion of foreground in all samples, as shown in Fig. \ref{h_1}. We can see that the foreground proportions of the majority of samples are less than 5\%. CNN-based methods usually do not perform well on these images with small and scattered objects. This further explains the poor results of CNN-based methods regarding metrics from Pr@0.5 to Pr@0.9. 

As to oIoU and mIoU scores, the performance gap between CNN-based methods and Transformer-based ones is much smaller. Nevertheless, the mIoU of our method can still outperform that of ConvLSTM by 16\%. We attribute this large improvement to three aspects: 1) our method leverages Swin Transformer \cite{liu2021swin} as the backbone, which is a stronger vision backbone compared with CNNs; 2) the self-attention mechanism used in Transformer networks could be helpful for segmenting small objects; 3) the designed LGCE module is more effective for learning more powerful multi-modal features. 

\subsubsection{Comparison with LAVT}
In this work, we propose an LGCE module to enhance shallow features by combining them with deep features under language guidance. To fairly validate the effectiveness of the LGCE module, we compare our method with LAVT by keeping other parts of the models the same. I.e., both models use ${BERT}_{BASE}$ as the language encoder and Swin-B as the visual backbone. As shown in Table \ref{exp_1}, our method outperforms LAVT in all evaluation metrics. On the one hand, our method demonstrates a significant improvement regarding Pr@0.6 and Pr@0.7 compared to LAVT. This means that there are more predictions with IoUs greater than 0.6 and 0.7. On the other hand, our method exhibits an approximate 2\% increase in mIoU, indicating a substantial advantage in detecting small objects. However, due to the strong ability of LAVT in detecting large objects, our method does not show notable enhancement in oIoU. Visualization results in Fig. \ref{visual1} exhibit that our method noticeably enhances the identification of small and scattered objects compared with LAVT.

\subsection{Ablation Study and Discussion}
LGCE aims to enhance visual representations by fusing multi-scale features under the guidance of linguistic features. Considering this motivation, the language guidance and cross-scale fusion are two core components of the LGCE module. To verify their effectiveness, we carry out ablation experiments and present numerical results in Table \ref{exp_2}. 

The first row represents experimental results of LAVT, which is adopted as our baseline. In the second row, we replace linguistic features with a randomly initialized token to remove language guidance. The clear performance drop indicates the importance of language guidance for multi-scale feature fusion in our model. In the third row, we maintain the remaining parts unchanged while eliminating the cross-scale fusion component. The results show that its segmentation performance in Pr@0.6, Pr@0.7, Pr@0.8, and Pr@0.9 decreases significantly compared with the full model. In the fourth row, the fully-fledged model clearly outperforms the baseline method and sets a new state-of-the-art performance on the RefSegRS dataset. The ablation studies can demonstrate the effectiveness of the proposed LGCE module. 

Regarding the computational complexity, we provide details on the floating point operations (FLOPs), the number of parameters (PARAMs), and multiply-accumulate operations (MACs) for both our model variations and the baseline model. As shown in Table \ref{exp_3}, our model incurs only a marginal increase in computational cost compared to the baseline method (LAVT), yet demonstrates significantly improved performance. While the model without cross-scale fusion has lower computational costs than the full model, it leads to a performance decline. In addition, the model without language guidance maintains the same network architecture as the full model, resulting in identical computational costs and the number of parameters.

\section{Conclusion}
\label{Conclusion}
Derived from the concept of referring natural image segmentation, we introduce RRSIS as a novel task within the domain of remote sensing. Considering the lack of datasets, we create a new dataset named RefSegRS. Specifically, we crop tiles from the SkyScapes dataset into images of $1200 \times 1200$ pixels using a sliding window with a stride of 600 pixels. Then, we generate linguistic expressions for each image with predefined templates. After manually filtering out uninformative image-language-label triplets, we ultimately obtain a dataset consisting of 4,420 image-language-label triplets across 285 scenes. To fully explore this task, we benchmark a variety of existing methods that are originally designed for natural images in computer vision and offer a comprehensive analysis and valuable insights. Our experimental results reveal that the direct adaptation of existing methods demonstrates limited efficacy in detecting small and scattered objects. Considering this, we propose a novel LGCE module to alleviate this problem. The effectiveness of our LGCE module in detecting such objects is verified by both quantitative and qualitative experimental results. For some challenging scenes characterized by cluttered and small objects, LGCE exhibits a remarkable enhancement, achieving an IoU improvement of more than 5\% in comparison to the baseline, as shown in Fig. \ref{visual0}.


\bibliographystyle{IEEEbib}
\bibliography{egbib}

\end{document}